\documentclass[conference]{IEEEtran}
\IEEEoverridecommandlockouts
\usepackage{cite}
\usepackage{amsmath,amssymb,amsfonts}
\usepackage{algorithmic}
\usepackage{graphicx}
\usepackage{textcomp}
\usepackage{xcolor}

\def\BibTeX{{\rm B\kern-.05em{\sc i\kern-.025em b}\kern-.08em
    T\kern-.1667em\lower.7ex\hbox{E}\kern-.125emX}}
\usepackage{enumitem}
\usepackage{booktabs}
\usepackage{bm}
\usepackage{gensymb}
\usepackage{ulem}
\usepackage[T1]{fontenc}    % use 8-bit T1 fonts
\usepackage{url}            % simple URL typesetting
\usepackage{multicol}
\usepackage[all]{nowidow}
\usepackage{tablefootnote}
\usepackage{threeparttable}
\usepackage{svg} % to import high-resolution vector graphics (PDF/SVG)
\usepackage{multirow}
\usepackage{array}
\usepackage{dblfloatfix}
\usepackage{comment}

\renewcommand\labelenumi{(\roman{enumi})}
\renewcommand\theenumi\labelenumi
\begin{document}

\title{Guiding Evolutionary Molecular Design: Adding Reinforcement Learning for Mutation Selection}

\author{\IEEEauthorblockN{Gaëlle MILON-HARNOIS}
\IEEEauthorblockA{\textit{Univ Angers, LERIA}, % \\
\textit{SFR MathSTIC}\\
F-49000 Angers, France \\
gaelle.milon-harnois@univ-angers.fr}
\and
\IEEEauthorblockN{Chaïmaâ TOUHAMI}
\IEEEauthorblockA{\textit{Univ Angers, LERIA}, % \\
\textit{SFR MathSTIC} \\
F-49000 Angers, France \\
ctouhami@etud.univ-angers.fr}
\and
\IEEEauthorblockN{Nicolas GUTOWSKI}
\IEEEauthorblockA{\textit{Univ Angers, LERIA}, % \\
\textit{SFR MathSTIC} \\
F-49000 Angers, France \\
nicolas.gutowski@univ-angers.fr}
\and
\IEEEauthorblockN{Benoit DA MOTA}
\IEEEauthorblockA{\textit{Univ Angers, LERIA}, % \\
\textit{SFR MathSTIC} \\
F-49000 Angers, France \\
benoit.damota@univ-angers.fr}
\and
\IEEEauthorblockN{Thomas CAUCHY}
\IEEEauthorblockA{\textit{Univ Angers, CNRS, MOLTECH-ANJOU}, % \\
\textit{SFR MATRIX}\\
F-49000 Angers, France \\
thomas.cauchy@univ-angers.fr}
}
\begin{comment}
\author{\IEEEauthorblockN{Gaëlle Milon-Harnois\textsuperscript{1}, Chaïmaâ Touhami\textsuperscript{1}, Nicolas Gutowski\textsuperscript{1}, Benoit Da Mota\textsuperscript{1}, Thomas Cauchy\textsuperscript{2}}
\IEEEauthorblockA{ \textit{ \textsuperscript{1}Univ Angers, LERIA, SFR MathSTIC, F-49000 Angers, France} \\
\textit{ \textsuperscript{2}Univ Angers, CNRS, MOLTECH-ANJOU, SFR MATRIX, F-49000 Angers, France}\\
\{gaelle.milon-harnois, nicolas.gutowski, benoit.damota, thomas.cauchy\}@univ-angers.fr}ctouhami@etud.univ-angers.fr
}
\end{comment}

\maketitle
\begin{abstract}
The efficient exploration of chemical space remains a central challenge, as many generative models still produce unstable or non-synthesizable compounds. To address these limitations, we present \textit{EvoMol-RL}, a significant extension of the \textit{EvoMol} evolutionary algorithm that integrates reinforcement learning to guide molecular mutations based on local structural context. By leveraging Extended Connectivity Fingerprints (ECFPs), \textit{EvoMol-RL} learns context-aware mutation policies that prioritize chemically plausible transformations. This approach significantly improves the generation of valid and realistic molecules, reducing the frequency of structural artifacts and enhancing optimization performance. The results demonstrate that \textit{EvoMol-RL} consistently outperforms its baseline in molecular pre-filtering realism. These results emphasize the effectiveness of combining reinforcement learning with molecular fingerprints to generate chemically relevant molecular structures.
\end{abstract}


\section{Introduction}
The chemical space of possible molecules is extraordinarily vast, rendering exhaustive exploration computationally intractable.  Theoretical estimates suggest that the number of drug-like molecules may exceed $10^{60}$\cite{bohacek_art_1996}, while even limiting the enumeration to small organic molecules with coherent structures can already yield over 160 billion structures \cite{ruddigkeit_enumeration_2012}. Yet, databases of experimentally known or commercially available molecules such as ZINC \cite{irwin_zinc_2012} or ChEMBL \cite{gaulton_chembl_2017} contain fewer than a billion compounds. This indicates an immense potential for discovery—spanning applications in drug development, materials science, and fundamental chemistry. Exploring this space to identify novel, synthetically feasible, and functionally relevant molecules is one of the central challenges in cheminformatics.

This combinatorial explosion poses fundamental obstacles for rational molecular discovery, especially when candidate evaluation is computationally expensive or experimentally costly. Accordingly, the past decade has seen the development of a wide variety of generative models aimed at the design of novel molecules optimized for target properties. These include: (1) \textbf{Rule-based methods}, notably template-based retrosynthetic planning \cite{segler_planning_2018}, expert-guided reactivity prediction \cite{coley_graph-convolutional_2019}, and extraction of graph-based reaction rules from large-scale reaction databases with atom-atom mapping \cite{phan_syntemp_2025}; (2) \textbf{Fragment-based approaches}, which assemble molecules by recombining known substructures, with junction-tree Variational Autoencoders as seminal examples \cite{jin_junction_2018}, and recent comprehensive reviews highlighting their strengths in ensuring chemical validity and interpretability \cite{voloboev_review_2024}; (3) \textbf{Evolutionary algorithms (EAs)}, which use iterative mutation and selection processes to explore chemical space, with growing interest in combining EA frameworks with synthesizability constraints \cite{jensen_graph-based_2019,leguy_evomol_2020, henault_chemical_2020}; (4) \textbf{Deep learning models} based on Textual representations such as the Simplified Molecular Input Line Entry System (SMILES) or graph representations, including autoencoders  \cite{gomez-bombarelli_automatic_2018}, graph neural networks, and diffusion-based generative models, as surveyed comprehensively in \cite{elton_deep_2019}; and (5) \textbf{Reinforcement learning (RL)} where agents iteratively construct or modify molecules based on sequential decision-making, has emerged as a powerful framework for molecular optimization. In chemistry, RL agents learn to construct or modify molecules by exploring chemical space with the aim of maximizing long-term rewards tied to molecular properties. Foundational work by Olivecrona \textit{et al.} \cite{marcus_olivecrona_molecular_2017} introduced policy-gradient-based SMILES generation, while Zhou \textit{et al.} \cite{zhou_optimization_2019} extended this to graph-based models using deep Q-networks. Recent reviews by Gow \textit{et al.} \cite{gow_review_2022} and Sridharan \textit{et al.} \cite{sridharan_deep_2024} provide comprehensive overviews of RL's theoretical foundations and practical implementations in chemistry, covering applications from molecule generation and retrosynthesis to geometry optimization. These works highlight the increasing relevance of RL as a general-purpose optimization tool in computational molecular science.

Despite significant advances, a persistent limitation across generative models is the production of unrealistic, unstable, or non-synthesizable molecules. This issue arises in various forms: random sampling in \textit{de novo} models often lacks chemical validity \cite{polykovskiy_molecular_2020}; latent space methods may suffer from overfitting, mode collapse, or poor inverse mapping \cite{renz_failure_2019}; fragment-based strategies can recombine motifs in sterically or electronically implausible ways without contextual coherence \cite{voloboev_review_2024}; and rule-based simulations may generate products outside synthetic feasibility due to incomplete encoding of reaction conditions or selectivity constraints  \cite{schwaller_mapping_2021}. These limitations typically stem from a lack of chemically informed constraints, insufficient local context during generation, or excessively permissive mutation and transformation rules \cite{elton_deep_2019}. 
In this landscape, \textit{EvoMol} offers a robust alternative by combining evolutionary search with chemically motivated filters, particularly those enforcing realistic atom environments and ring systems \cite{cauchy_definition_2023}. However, like most evolutionary algorithms, it applies stochastic mutations that are agnostic to the local molecular context \cite{henault_chemical_2020}. As a result, many generated structures remain chemically irrelevant and must still be evaluated and filtered out. This inefficiency becomes particularly problematic in scenarios where molecular evaluation is computationally expensive (e.g., docking, quantum chemistry) or experimentally limited \cite{bannwarth_extended_2021}.

To overcome the limitations of context-agnostic mutations in evolutionary algorithms, we introduce \textit{EvoMol-RL}, an extension of the \textit{EvoMol} framework that integrates \textit{reinforcement learning} (RL) to guide the molecular mutation process. By incorporating local molecular environments encoded via Extended Connectivity Fingerprints (ECFPs) \cite{rogers_extended-connectivity_2010}, \textit{EvoMol-RL} learns context-aware mutation policies that favor chemically meaningful and synthetically plausible transformations. This approach combines the stochastic exploration strength of evolutionary algorithms with the policy learning capabilities of RL, thereby improving both the efficiency and realism of chemical space navigation.
The main contributions of this work are as follows:
(1) we enhance \textit{EvoMol} with a reinforcement learning agent that selects mutations conditioned on the molecular context (\textit{EvoMol-RL});  
(2) we introduce a dynamic and adaptive action space, structured around ECFP-derived features, which restricts chemically invalid or unproductive mutations;  
(3) we show that \textit{EvoMol-RL} significantly outperforms the original \textit{EvoMol} in terms of chemical validity, synthesizability, and property-based optimization performance.

The paper is organized as follows. Section~\ref{sec:BackMot} reviews relevant background on molecular generation and policy learning. Section~\ref{sec:PbMet} presents the problem formulation and describes the \textit{EvoMol-RL} architecture in detail. Section~\ref{sec:Results} reports experimental results and comparative benchmarks. Section~\ref{Conclusion} concludes with key findings and outlines directions for future research.

\section{Background}\label{sec:BackMot}

\subsection{Molecular Basics}\label{ssec:MolBasic}
A molecule is a discrete chemical entity composed of atoms held together by covalent bonds. Atoms are characterized by their atomic number or symbol and valence—defined as the number of covalent bonds they can form—while bonds may vary in order (single, double, triple). Molecular design efforts in organic chemistry typically focus on a subset of "heavy" atoms, including carbon (C), nitrogen (N), oxygen (O), fluorine (F), sulfur (S), phosphorus (P), chlorine (Cl), and bromine (Br), in addition to hydrogen (H).

\subsection{Molecular Representation}\label{ssec:mrep}
A molecule can be represented in various formats designed to be compact, reproducible, unambiguous, flexible, and suitable for computational analysis. Among these, two are particularly popular.

\paragraph{Molecular graphs} 
they form the backbone of many cheminformatics approaches. 
Atoms, except H, are represented as vertices with atomic symbol, and bonds as edges labeled by bond types \cite{leguyPhD2022}. Hydrogen atoms are usually implicit and inferred from standard valence rules.  Mathematically, a molecular graph can be defined as in equation \eqref{eqGraph}.
\begin{equation}
    G = (V, E, f_{\alpha\tau},f_\beta,f_\gamma)
    \label{eqGraph}
\end{equation}
where $V$ is a set of vertices representing the atoms of the molecule;  
$E \subseteq V \times V$ is a set of edges representing the chemical bonds between atoms;
$f_{\alpha\tau}: V \xrightarrow{} \{C, N, O, F, P, S, Cl, Br\}$ is the vertex labeling function, associating each vertex with an atomic symbol;
$f_\beta: E \xrightarrow{} \{1,2,3\}$ is the edge labeling function, associating each edge with a bond type;
$f_\gamma: V \xrightarrow{} \mathbb{Z}$ is the formal charge labeling function, associating a formal charge to each vertex. 

While this discrete representation is well-suited to computation, challenges remain: exact graph matching (isomorphism) is a NP-complete decision problem, and comparing graphs via edit distance is a NP-hard optimization problem. Nevertheless, chemists have developed rules (e.g., canonicalization procedures) to provide consistent and unique graph traversals in most common cases. 

Classical molecular drawings have some additional particularities: vertices without letters correspond to carbon atoms, and hydrogens bonded to carbons are implicit, while other hydrogens are explicit. Fig.~\ref{fig:ecfp} presents on the left, an example of the graph representation for the acetylsalicylic acid molecule.

\paragraph{Textual Representations} 
Textual representations such as the SMILES provide a linear encoding of molecular graphs with canonical variants, allowing string-based comparisons \cite{weininger_smiles_1988}. Among the drawbacks of this representation, SMILES discrete syntax does not always align with chemical similarity, and edit distances such as Levenshtein may fail to capture structural features. The top left of fig.~\ref{fig:ecfp} shows an example of the SMILES representation for the acetylsalicylic acid molecule.

\subsection{Molecular Context}\label{ssec:mcontx}
\begin{figure*}[!t]
    \centering
    \includegraphics[width=\linewidth, trim=2cm 11cm 2cm 2cm, clip=true]
    {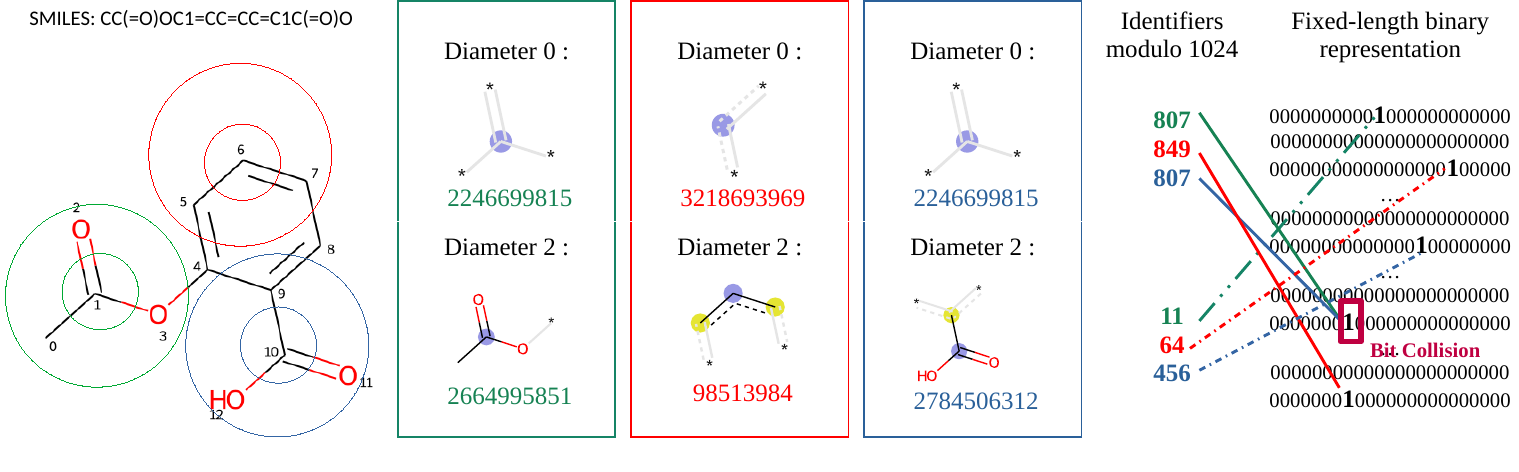}
    \caption{ECFP generation process on Acetylsalicylic acid molecule and generation of the fixed-length fingerprint. \textit{Blue circles correspond to the central atoms. 
    }    }
    \label{fig:ecfp}
\end{figure*}

The graph-based approach is the foundation for defining structural descriptors like Extended Connectivity FingerPrints (ECFP) \cite{rogers_extended-connectivity_2010}. ECFPs are topological, circular fingerprints that iteratively encode substructures around each atom in a molecule, capturing local environments within a specified diameter (see Fig.~\ref{fig:ecfp}). 

At diameter $d=0$ (noted ECFP$_0$), only intrinsic atomic features—such as atomic number, formal charge, number of bonded heavy atoms, and number of attached hydrogens—are considered. These features are hashed into unique integer identifiers. At $d=2$ (ECFP$_2$), the process incorporates first-neighbor environments; for each atom, the hashed identifiers of its immediate neighbors are concatenated with its own, then re-hashed. This iterative procedure continues until a target diameter is reached (typically $d=4$; ECFP$_4$). The resulting identifiers describe unique atom-centered environments across the molecule. These are collected and folded (e.g., modulo 1024 or 2048) into a fixed-length binary vector that encodes the presence (1) or absence (0) of specific environments. In our work, ECFPs are used to define the molecular context on which mutation decisions are conditioned.

\subsection{Unrealistic environments: Silly Walks}\label{ssec:Silly}
\textit{Silly Walks} (SW) is a computational metric that quantifies structural implausibility (or molecular abnormality) on a scale from 0, the least \textit{silly} molecule, to 1, the most \textit{silly} one. Following the implementation available in Patrick Walters' GitHub repository \cite{walters_silly_walks_2022}, the metric evaluates the proportion of \textit{silly} bits in a molecule, i.e. the ECFP$_d$ (with $d \in \{0, 2, 4\}$) that never appear in a reference dataset, usually, the ChEMBL \cite{buhlmann_chembl-likeness_2020} or Zinc \cite{irwin_zinc_2012} databases.

\subsection{The \textit{EvoMol} Algorithm}\label{ssec:EvoMol}
In the field of de novo generation of molecules, \textit{EvoMol} is an evolutionary algorithm that optimizes an arbitrary objective function ($OF$) by performing atomic level mutation on a set of molecules \cite{leguy_evomol_2020}. 
At each step, the algorithm selects up to ten best molecules based on this $OF$. Each selected molecule is then processed individually: random actions are applied to explore its neighborhood to find a better solution. For each molecule, a maximum of 50 mutations are attempted to find an improver. 

In this study, among all \textit{EvoMol}'s actions, we will only consider the most elementary ones, which alone allow us to cover the entire molecular space.
Three primary atomic level actions are available: adding an atom (\textit{AddA}), removing an atom (\textit{RmA}) and changing bonds between 2 atoms (\textit{ChB}), which includes creation and deletion of a bond.

\begin{figure*}[!t]
    \centering
    \includegraphics[width=\linewidth]
{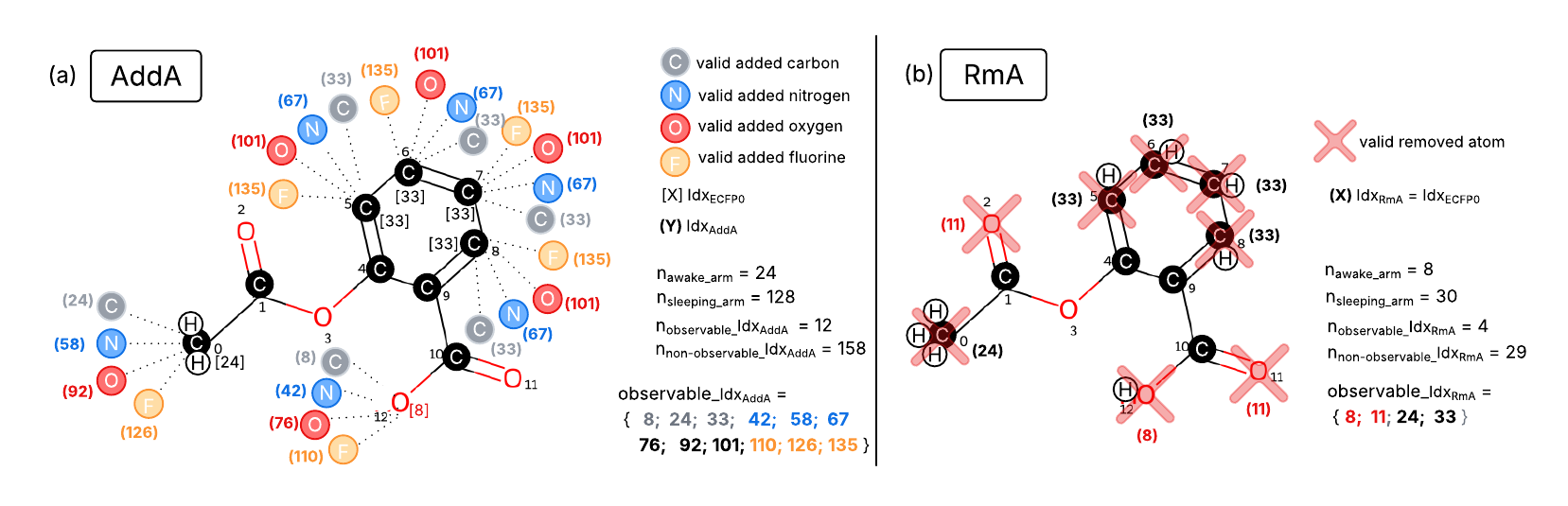}
    \caption{Valid mutations starting from the acetylsalicylic acid molecule and corresponding $Idx_{ECFP_0}$ and $Idx_a$. (a) \textit{AddA} valid mutations with (C, N, O, F) candidate atom set. (b) \textit{RmA} valid mutations. 
    }
    \label{fig:Valid_actions} 
\end{figure*}

During each iteration, \textit{EvoMol} systematically evaluates the validity of every available action ($a\in \mathcal{A}$) for each vertex in the molecular graph, where $\mathcal{A}$ is the set of potential actions defined by the user within \(\left\{AddA, RmA, ChB \right\}\).
Therefore, all actions are not valid for each vertex as illustrated by Fig.~\ref{fig:Valid_actions} which presents all the \textit{AddA} and \textit{RmA} valid mutations on acetylsalicylic acid molecule. For the sake of space and clarity, valid \textit{ChB} mutations are not reported.
The valid mutations list ($Valid_{(a, idx_p)}$) 
contains all combinations of an action $a\in\mathcal{A}$ paired with an integer $idx_p$ that denote the positional index of each valid $a$ within the molecular structure, optionally crossed with additional parameters such as atom to add for \textit{AddA} or type of new bond for \textit{ChB}. 
Selection of a mutation $(a, idx_p)$ is then performed through random sampling from $Valid_{(a, idx_p)}$.

Following the execution of the selected mutation on the molecule to improve, filters are implemented to ensure that the resulting molecule does not contain any \textit{silly} bits (see Sec.~\ref{ssec:Silly}) and represents a genuinely novel compound that is not present within the existing molecular population.
When a generated molecule successfully passes the filters, the $OF$ is evaluated and compared to that of the parent molecule. Only new molecules showing an improved $OF$
relative to their initial counterpart are inserted into the pool of candidates. 
This iterative process continues until a stopping criterion is met, usually a predetermined number of steps. This depiction of \textit{EvoMol} is simplified in order to focus on the elements necessary in our study, more details are available in the article describing the method \cite{leguy_evomol_2020}.

\subsection{Reinforcement Learning and Sleeping Bandit Problem}\label{ssec:RL}
Since \textit{EvoMol-RL} relies on both a probability matching like method and the adversarial sleeping bandit problem, this section reminds all the related key concepts that underlie our approach. 

\paragraph{Multi-armed bandits}
The k-armed bandit \cite{lattimore2020bandit} is a reinforcement learning \cite{sutton1998reinforcement} instance with a single state and a set of \(k\) independent actions (“arms”) $i\in k$. Each arm is associated with a fixed reward distribution of mean $\mu_{i} \in [0,1]$. The vector of expected rewards is denoted by $D_r = (\mu_{1}, \ldots, \mu_{k}) \in [0,1]^k$. The problem is sequential: let $T \in \mathbb{N}^*$ be the time horizon, at each iteration $t \in {1, \ldots, T}$, the learner chooses an arm \(i_t\in K\) and receives a loss \(\ell_{t,i_t}\in[0,1]\) (or reward \(r_{t,i_t}=1-\ell_{t,i_t}\)) (i.e. a reward $r_{i_t} \in \left\{0,1\right\}$ is drawn from a Bernoulli distribution with parameter $\mu_{i}$). The goal is to maximize the cumulative reward (or minimize the regret) by learning to favor arms with higher expected returns.

Its performance is thus measured by the (pseudo-)regret defined in equation \eqref{eqRegret}.
\begin{equation}
\text{Reg}_T
= \sum_{t=1}^T \ell_{t,i_t}
- \min_{i\in[K]}\sum_{t=1}^T \ell_{t,i}.
    \label{eqRegret}
\end{equation}

\paragraph{Adversarial sleeping bandits}
In the adversarial sleeping bandit (ASB) \cite{kleinberg2019bandits} and \cite{combes2015combinatorial}, losses \(\{\ell_{t,i}\}_{i = 1}^k\) may be chosen by an adversary and only a subset \(K_t\subseteq K\) of “available” arms is offered at each round.  
The learner must pick \(i_t\in K_t\).  
As defined in equation \eqref{eqRegretAw}, regret is computed against the best fixed arm that was "awake" whenever it was needed:
\begin{equation}
\text{Reg}_T^{\text{sleep}}
= \max_{i\in[K]}
\sum_{t=1}^T \bigl(\mathbb{I}[i\in K_t]\cdot \ell_{t,i_t}
- \ell_{t,i}\bigr).
    \label{eqRegretAw}
\end{equation}
Note that classical algorithms like EXP3-IX guarantee \(\widetilde{O}(\sqrt{T})\) regret \cite{neu2015explore}.

\paragraph{Connection with \textit{EvoMol-RL}}
In \textit{EvoMol-RL}, each chemical mutation is modeled as an arm.  
Because the set of valid mutations depends on the current molecular graph, arms naturally sleep.  
Moreover, the reward acceptance by the SW filter is non-stochastic and potentially adversarial, motivating the ASB formulation.  
Our empirical probability matching heuristic strategy (See Sec.~\ref{ssec:GPS}) is a roulette wheel implementation adapted to the huge, sparse contextual state space induced by ECFP contexts (see Sec.~\ref{ssec:MolContext}).  
This formal link justifies the exploration-exploitation machinery introduced later.

\subsection{Motivation}\label{ssec:motivation}
Since \textit{EvoMol} is context-blind, it performs mutations randomly on the molecular graph. Without feedback on which mutations are efficient and which are ineffective, the agent lacks a reference point to decide which arm to select. We hypothesize that mutation selection can be improved through a heuristic based on RL that leverages molecular context using ECFP for learning. Since not all mutations and not all ECFP$_d$ indices are available at each step, and rewards (consisting in success counts) are computed iteratively, the adversarial sleeping bandit framework appears well-suited to adress this problem. Learning is guided by a reward function based on the molecule's ability to pass the \textit{SW} filters.

\section{Problem Setting}\label{sec:PbMet}
In this section, we state our problem and describe our method: \textit{EvoMol-RL}, \textit{EvoMol} enhanced with Reinforcement Learning. 

\subsection{Problem setup}\label{ssec:GPS}
At each iteration, the algorithm operates under a probability matching adversarial sleeping bandit-inspired heuristic approach, where the set of valid mutations (arms) $\mathcal{M}_t$ at time step $t \in [1, T]$ varies dynamically depending on the molecular graph structure. 

Namely, let $T \in \mathbb{N}^{*}$ be the time horizon. Let $\mathcal{M} =\{m_1,...,m_{k_T}\} \subseteq \Omega$ be a given set of $k_T $ available independent valid mutations encountered during an entire \textit{EvoMol} molecular generation process. $\Omega$ represents the whole universe of possible mutations that can be encountered during any molecular generation process. At each time step $t\in [1, T]$, a subset of $k_t$ distinct valid mutations $\mathcal{M}_t  =\{m_1^{(t)},...,m_{k_t}^{(t)}\} \subseteq \mathcal{M} \subseteq \Omega$ is available, which depends on the current state of the molecular graph $G_t$, thus resulting in non-stationary and context-dependent dynamics.
%Upon receiving the valid mutations set $\mathcal{M} _t$, the optimal action $m^*$ must be determined. 
Each valid mutation $m_j \in \mathcal{M}_t$ ($j \in \{1,..., k_t\}$) is characterized by the action $a$ applied at positional index $idx_p$ on $G_t$, leading to a local context $c$ derived from an ECFP-like molecular encoding.
Let $p_{m_j}^{(t)}$ denote the empirical success probability for mutation $m_j$ at time t,  computed as the success-to-trial ratio for action $a$ within the local context $c$. A stochastic policy $\pi_t(m) \propto p_{m_j}^{(t)}$ is then employed to select an action $m_t \in \mathcal{M}_t$, applying a dynamic proportional selection mechanism similar to a roulette wheel sampling strategy. This method inspired from probability matching relies on $p_{m_j}$ build upon past observations. Upon executing mutation $m_t$, a reward $r_{t, m_t} =1-\ell_{t,m_t}$ is observed based on the \textit{SW} metric (see Sec.~\ref{ssec:Silly} and Table~\ref{Table:NotationASB}). Finally, $p_{(a,c)}$ for the selected action $a$ within the selected local context $c$, is  updated according to the observed binary reward $r_{t, m_t} \in \{0,1\}$. The detailed update mechanism for $p_{(a,c)}$ (denoted $p_{Idx_{a}(c)}$) is presented in the next section ~\ref{ssec:MolReal}.

\begin{table}[!b]
\centering
\caption{Notation used in the Sleeping Bandit formulation of \textit{EvoMol-RL.}}\label{Table:NotationASB}
\begin{tabular}{@{}ll@{}}
\toprule
\textbf{Symbol} & \textbf{Meaning} \\
\midrule
\(\Omega\) & Whole universe of distinct mutations\\
\(k\) & Total number of distinct mutations\\
\(\mathcal{M}\) & Set of distinct mutations encountered\\
\(k_T\) & Total number of distinct mutations encountered \\
\(\mathcal{M}_t\) & Set of mutations valid (“awake”) at step \(t\) \\
\(k_t\) & Number of distinct mutations valid (“awake”) at step \(t\) \\
\(m_t\) & Mutation actually applied at step \(t\) \\
\(\ell_{t,m_t}\) & Loss of  $m_t$ at step $t$
(1 if mutant fails SW filter, 0 otherwise) \\
\(\text{Reg}_T^{\text{sleep}}\) & Adversarial sleeping-bandit regret up to step \(T\) \\
\bottomrule
\end{tabular}
\end{table}

According to this problem setup, the subsequent subsections will examine the key point formalization of molecular context $c$, describe the concept of sleeping mutations, and explicit the evaluation of molecule realism.

\smallskip
\subsubsection{Molecular Context} \label{ssec:MolContext}
For each action, the molecular context must encode both the ECFP$_d$ identifier ($ID_{ECFP_d}$) and, when applicable, the specific option index for actions with multiple variants. Specifically, to fully characterize the contextual state space, \textit{AddA} contexts require the consideration of the index of the atom type to be added ($Idx_{\alpha\tau}$) within the candidate atom set, while \textit{ChB} contexts require the bond type index ($Idx_{\beta}$). 
Because the number of possible contexts varies across different action types, molecular context analysis must be performed on an action-specific basis.

The function $Idx_{a}(c)$ generates a unique identifier for each known and realistic context $c$ of each action $a$. It is computed from the positional index of $ID_{ECFP_d}$ within the valid ECFP$_d$ list ($Idx_{ECFP_d}$) combined with either $Idx_{\alpha\tau}$ for \textit{AddA} or $Idx_{\beta}$ for \textit{ChB}. 
Fig.~\ref{fig:Valid_actions} illustrates this encoding scheme using acetylsalicylic acid molecular structure, with diameter 0 ECFP list containing 33 $ID_{ECFP_0}$. Fig.~\ref{fig:Valid_actions} shows the $Idx_{ECFP_0}$ values and the corresponding $Idx_{AddA}$ and $Idx_{RmA}$ indices calculated for all \textit{AddA} and \textit{RmA} valid mutations applicable to this molecular system. For example, carbon atoms where \textit{AddA} operations are valid (Fig.~\ref{fig:Valid_actions}a) correspond to two distinct ECFP$_0$ environments ($Idx_{ECFP_0}$ = 24 and 33). Since four atom types (C, N, O, F) can be added to each environment, this results in 8 different $Idx_{AddA}$ values (24, 33, 58, 67, 92, 101, 126, 135). Similarly, the oxygen atoms where \textit{AddA} operations are valid exhibit only one $Idx_{ECFP_0}$ value (8), resulting in 4 potential $Idx_{AddA}$ values (8, 42, 76, 110). In total, 12 $Idx_{AddA}$ values are valid for step 1.
An inverse function applied to these $Idx_{a}$ enables the recovery of the contextual $ID_{ECFP_0}$.
Finaly, as shown, identical index values may occur multiple times within a single algorithmic iteration.

\smallskip
\subsubsection{Dynamic mutations set and sleeping arms} \label{ssec:SA}
Since the molecular graph undergoes structural modifications across successive steps, the valid mutations set exhibits dynamic variations that correspond to the evolving configuration of atoms and bonds within the molecular graph. 
For example, as shown in Fig.~\ref{fig:Valid_actions}a, the carbon atom at position 1 of the acetylsalicylic acid molecule has already formed 4 bonds and therefore cannot undergo any \textit{AddA} mutation. The 4 theoretical arms (namely, the addition of carbon, nitrogen, oxygen, and fluorine atoms at position 1) therefore remain sleeping. Similarly, the removal of the same atom (Fig.~\ref{fig:Valid_actions}b) would necessarily result in molecular cleavage into two separate entities; consequently, the mutation $(RmA, 1)$ is sleeping at this step.
Based on the total mutations frequency calculated from the default maximum heavy atom count per molecule (38), each action exhibits stepwise variation in sleeping and awake mutations frequencies, with sleeping mutations theoretically predominating (refer to $n_{awake-arm}$ and $n_{sleeping-arm}$ in both panels of Fig.~\ref{fig:Valid_actions}).

Given that molecular contexts are intrinsically linked to mutations, the set of observable $ID_{ECFP_d}$ varies accordingly, rendering the collection of observable mutation contexts inherently non-stationary in nature. At any given time step, the vast majority of theoretical $Idx_{a}(c)$ remains unavailable and is therefore classified as non-observable $ID_{ECFP_0}$, as demonstrated by the comparison of observable and non-observable $Idx_{a}$ frequencies in Fig.~\ref{fig:Valid_actions}.
For instance, Fig.~\ref{fig:Valid_actions} shows that the acetylsalicylic acid molecule lacks both F and N atoms. Consequently, all $ID_{ECFP_0}$ related to these atoms remain inactive at step 1. If mutation chosen adds a N atom at position 12 (Fig.~\ref{fig:Valid_actions}a), $ID_{ECFP_0}$ related to this atom will be activated at Step 2. Since the O atom at position 12 no longer maintains a bond with hydrogen, the \textit{AddA} action on this atom will become inactive at this step and related context indices 8, 76, and 110 will consequently become non-observable $Idx_{AddA}$.

\smallskip
\subsubsection{Evaluation of Molecule Realism and Probability Matching } \label{ssec:MolReal}
In \ref{ssec:Silly}, \textit{SW} metric was introduced as a measure for evaluating the proportion of \textit{silly} bits in molecular representations. In this study, \textit{SW} serve a dual analytical purpose both detailed in this subsection: they are employed both as a criterion for selecting the most relevant actions during the molecular generation process and as a comparative metric for evaluating the realism of molecules generated by \textit{EvoMol-RL} against those produced by the standard \textit{EvoMol} approach. 

At each step, mutation $m_j$ (tuple ($a$, $idx_p$)) selection is probability-matching based, using empirical success rates of corresponding $Idx_{a}(c)$ to pass the \textit{SW} filter as implicit rewards, defined in equation \eqref{SuccessRate}. let 
$n_{Use}(Idx_{a}(c))$ be the frequency of $Idx_{a}(c)$ selection throughout the process and $n_{Success}(Idx_{a}(c))$, its frequency to generate a new molecule passing \textit{SW} filter (without \textit{silly} bit), $Idx_{a}(c)$ success rate, $p_{Idx_{a}(c)}$ is computed as follow: 

\begin{equation}
    p_{Idx_{a}(c)} = \frac{n_{Success}(Idx_{a}(c))}{n_{Uses}(Idx_{a}(c))}
    \label{SuccessRate}
\end{equation}

For each valid mutation $m_j \in \mathcal{M}_t$ ($j \in \{1,..., k_t\}$) (tuple $(a, idx_p)$ in $Valid_{(a, idx_p)}$); $m_j$ success rate, $p_{m_j}$, is the success rate of the corresponding action context $p_{Idx_{a}(c)}$. To provide opportunities for unselected contexts, a minimum success rate, denoted as $p_{min}$, is initially defined by the user. Subsequently, to ensure data smoothing, valid mutations success rates undergo normalization according to equation \eqref{NormalizedSuccessRate}:
\begin{equation}
    w_{m_j} = \frac{max(p_{min}, p_{m_j})}{\sum\limits_{m_j \in \mathcal{M}_t; j \in \left\{1, ..., k_t\right\}} max(p_{min}, p_{m_j})}
    \label{NormalizedSuccessRate}
\end{equation}
The list of weights $w_{m_j}$ is then used to perform weighted randomization of a mutation $(a, idx_p)$. This weighting scheme enables probabilistic selection where each mutation's likelihood of being chosen is proportional to its corresponding weight value, thereby implementing a stochastic sampling mechanism that favors tuples with higher associated weights while still maintaining the possibility of selecting those with lower weights.

By incorporating existing $ID_{ECFP_d}$ within $Idx_{a}(c)$ and leveraging their empirical efficacy in generating chemically realistic molecules, the proposed method is expected to reduce the number of newly generated molecules rejected by the \textit{SW }filter  compared to those produced by \textit{EvoMol} algorithm.

\smallskip
\subsection{Method to balance exploration and exploitation}
The exploitation-exploration trade-off balances using known high-reward actions versus testing new potentially better actions. Three strategies were tested in this protocol.
Constant strategy maintains the same exploration rate $\varepsilon$ throughout learning to randomly explore, with low values favoring exploitation and high values promoting exploration. Moderate  $\varepsilon$ values (e.g., 0.1, 0.2) provide balanced exploration-exploitation trade-off whereas higher values (from 0.3) prioritize exploration with frequent random action selection.
Epsilon-greedy and Power Law ($PL$) strategies start with high exploration ($\varepsilon_0 = 1$) that gradually decrease over time, allowing initial broad search followed by focused exploitation. In formulation of exploration rate following a power law distribution \eqref{eq: epsPL}; $\alpha$ value determine how rapidly exploration decreases over time. This function is done by $\lambda$ in \eqref{eq: epsGreedy} equation.
\begin{equation}
    \varepsilon_{Greedy}(t) = max(\varepsilon; \varepsilon_0 . e^{-\lambda.t})
    \label{eq: epsGreedy}
\end{equation}
\begin{equation}
    \varepsilon_{PL}(t) = max(\varepsilon; \frac{\varepsilon_0}{(1+t)^\alpha})
    \label{eq: epsPL}
\end{equation}
At each step $t$, exploration versus exploitation is determined by comparing the current epsilon value, $\varepsilon_{curr}$ (computed with \eqref{eq: epsGreedy} for Epsilon-greedy approach or \eqref{eq: epsPL} for PL strategy) against a random value $alea \in [0; 1]$.

These approaches were tested with different values for $\varepsilon$; $\alpha$ and $\lambda$ parameters to determine optimal balance between immediate rewards and long-term performance discovery.

\subsection{\textit{EvoMol-RL} algorithm} \label{ssec:EvoRL} 
\textit{EvoMol-RL} algorithm uses the \textit{EvoMol} algorithm as detailed in \cite{leguy_evomol_2020} (source code available at https://github.com/jules-leguy/EvoMol). The primary modification involves replacing the SearchNeighbour function with the following implementation\footnote{The comprehensive Python program is hosted at https://github.com/gmilha/EvoMol\_ICTAI}: 

\bigskip
\boxed{\begin{array}{l}
%\begin{empheq}[box=\fbox]{equation*}
\text{Compute  $Valid_{(a, idx_p)}$ in $CurrMol$ (Sec. \ref{ssec:EvoMol})}\\
\mathcal{M}_t = Valid_{(a, idx_p)} \\
m_{t} = \left\{\begin{array}{l}
                \textbf{if } \text{random}(alea) > \varepsilon_{Curr} \text{ \eqref{eq: epsGreedy} \eqref{eq: epsPL}:}\\                
                    \text{random}(m_j) \in \mathcal{M}_t                    \\
                    \textbf{else for} \text{ each } m_j \in \mathcal{M}_t \text{ ; } j \in {1, ..., k_t}:
                    \\
                    \left\{\begin{array}{l}
                            \text{Get } Idx_{a}(c)\\
                            \text{Compute } p_{Idx_{a}(c)} \text{ \eqref{SuccessRate}} \\
                            \text{Compute } w_{m_j} \text{ \eqref{NormalizedSuccessRate}}\\ 
                            \end{array}
                    \right. \\
                \text{   Weighted random}(m_j) \in \mathcal{M}_t
                \end{array}
            \right.\\
\text{Apply $m_{t}$ to $CurrMol$ generating $NewMol$ }\\
\text{Increment $n_{Uses}({Idx_{a}(c)_{t}})$}
%\end{empheq}
\end{array}
}

\bigskip
\textit{EvoMol-RL} algorithm subsequently verifies whether the $NewMol$ satisfies the filtering criteria and observes the reward, $r_{t, m_t}$, which depends on whether $NewMol$ represents an OF improvement. The success frequency of $Idx_{a}(c)_{t}$ is then updated according to the definition provided in \eqref{eq: nSucc}.
\begin{equation}
  n_{Success}({Idx_{a}(c)_{t}}) = n_{Success}({Idx_{a}(c)_{t}}) + r_{t, m_t}
  \label{eq: nSucc}
\end{equation}

\section{Empirical Evaluation}\label{sec:Results}

\subsection{Experimental Protocol}\label{ssec:XP}
To systematically evaluate the impact of key algorithmic parameters, an exhaustive enumeration of hyperparameters combinations was performed over ECFP diameter $d \in \left\{0, 2\right\}$, $\varepsilon \in \left\{0.1, 0.2, 0.3\right\}$, and exploration strategies: epsilon-greedy ($\lambda \in  \left\{10^{-3},10^{-2},10^{-1}\right\}$), Power Law ($\alpha \in [0.25, 0.4]$ in steps of 0.05) and constant epsilon value fixed at the corresponding $\varepsilon$ throughout the run. 

A series of ten experimental runs per configuration was conducted using predetermined constant seeds to ensure reproducible comparisons between different script configurations. 
Each run, started from acetylsalicylic acid molecule, was executed for a duration of 500 steps, based on the assumption of fast model convergence, with the \textit{SW}
objective function serving as the evaluation metric. 
Only one action is performed between each evaluation to avoid having to deal with delayed rewards.
The optimization process involves to maximize the occurrence of realistic molecules (those passing the \textit{SW} filter). Primary criterion is the rate of generated molecules passing the \textit{SW} filter throughout the run, \textit{i.e.} percentage of realism.
Secondary criterion is the rate of generated molecules not already in population P, \textit{i.e.} percentage of novelty. 
For all configurations, both criteria obtained with ECFP$_0$ and ECFP$_2$ are compared to actual \textit{EvoMol} results.

We would like to emphasize that by initializing the population with the same single molecule and by not activating any of \textit{EvoMol}'s specific parameters to favor the diversity of solutions, we are deliberately setting the algorithm in such a way that the two evaluated criteria are only influenced by the choice of mutations. This special case allows us to better assess the impact of our method. We therefore expect to observe an increase in realism, the objective function, at the cost of a decrease in novelty.

Among all tested parameters, only $\varepsilon$ meaningfully influences realism: variability remains 1\% regardless of exploration method, $\lambda$, or $\alpha$. Consequently, Table~\ref{tab: Results} and results analysis section (\ref{ssec:RA}) only shows the best-performing configuration for each ($\varepsilon$, ECFP diameter) pair. The peak realism is achieved by Power-Law ($\alpha = 0.35$) and Greedy ($\lambda = 0.10$). Moreover, because Power-Law is slightly faster and more stable, it serves as the reference for reporting means and inter-run standard deviations.

\subsection{Results Analysis}\label{ssec:RA}
Over the 10 runs for each method we compute and report in Table~\ref{tab: Results}: 1)~The realism percentage mean of generated molecules (before filtering) along with its standard deviation; 2)~The novelty percentage mean along with its standard deviation.

\begin{table}[tbp]
\caption{Percentage realism and percentage novelty means and standard deviations of molecules generated by \textit{EvoMol} and \textit{EvoMol-RL}.}
\begin{center}
\scalebox{0.95}{
\begin{tabular}{ | c | c | c | c| c | }
\hline
\textbf{$\varepsilon$} & \multicolumn{2}{|c|}{\textbf{\% realism mean $\pm$ std }} & \multicolumn{2}{|c|}{\textbf{\% novelty mean $\pm$ std }}\\
\hline	
  \textbf{\textit{EvoMol}} & \multicolumn{2}{|c|}{$  0.51 \pm 0.01$ } & \multicolumn{2}{|c|}{$0.58 \pm 0.01$ }\\
  \hline
  \textbf{\textit{EvoMol-RL}} & \textbf{\textit{ECFP$_0$}} & \textbf{\textit{ECFP$_2$}} &\textbf{\textit{ECFP$_0$}} & \textbf{\textit{ECFP$_2$}} \\
    \hline
 0.1 &  $ 0.78 \pm 0.01$ & $ 0.82 \pm 0.01$ &
 $0.36 \pm 0.01$ & $ 0.32 \pm 0.01 $ \\
\hline
 0.2 & $ 0.73 \pm 0.01$ & $ 0.76 \pm 0.01 $ & 
 $ 0.41 \pm 0.01 $   & $ 0.38 \pm 0.01 $ \\
  \hline
0.3 & $ 0.68 \pm 0.01$  & $0.69 \pm 0.01 $ & $
 0.46 \pm 0.01 $ & $0.44 \pm 0.01$\\
 \hline
\end{tabular} \label{tab: Results}
}
\end{center}
\end{table}

Despite the limited structural information provided by ECFP$_0$, a substantial increase in pre-filtering realism percentage is observed, achieving 17$\%$ ($\varepsilon = 0.3$) to 27$\%$ ($\varepsilon = 0.1$) higher realism relative to \textit{EvoMol}. When the richer ECFP$_2$ fingerprint is used instead, the best configuration reaches an average of 82$\%$ realistic molecules for $\varepsilon = 0.1$. 
For this optimal $\varepsilon$ configuration, a \textit{Kruskal-Wallis} test is subsequently conducted to highlight inequalities among the mean realism percentages across \textit{EvoMol}, \textit{EvoMol-RL} ECFP$_0$, and ECFP$_2$. %i.e. we test the \textit{null hypothesis $H_0$: "There is no significant difference between the methods' results"}
Herein, the \textit{Kruskal-Wallis} test demonstrates significant statistical differences in realism performance between the three methods under investigation ($p$-value $ < 0.001$).
%, $H_0$ is rejected).
Thus, a \textit{Wilcoxon} signed rank tests is performed over the realism to determine whether the results between each pair of methods differ significantly
%i.e. we test the \textit{null hypothesis $H_0$: "There is no significant difference between the results of each pair of methods"}
. The significance will be further given with the $p$-value.

%%%%%%% Figure placée ici pour qu'elle soit située en bas de la page où elle est appelée. %%%%%%%%%%%%%
\begin{figure}[!b]
    \centering
    %[trim={left bottom right top},clip]
    \includegraphics[trim={6cm 4.7cm 5.4cm 4.6cm},clip, width=\linewidth]%[width=3.7in]%[width=\linewidth]
{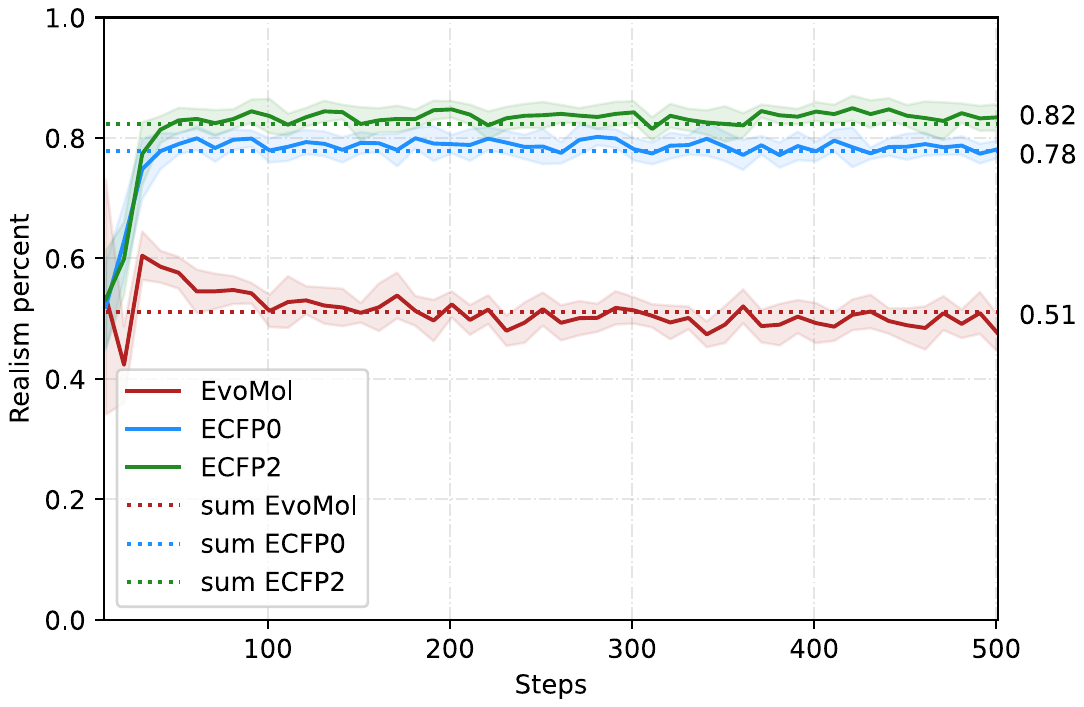}
    \caption{
    Percentage of realistic molecules generated across testing steps from the Power Law configuration with parameters $\alpha$=0.35 and $\varepsilon$=0.1. The solid line represents the mean values, with standard deviations indicated by the highlighted regions, computed using a sliding window of 10 steps. Results obtained with \textit{EvoMol} and \textit{EvoMol-RL} with consideration of ECFP$_0$ or ECFP$_2$ are respectively depicted in red, blue and green. 
    }
    \label{fig:PctRealismstep} 
\end{figure}

As illustrated in Table~\ref{tab: Results} and in Fig.~\ref{fig:PctRealismstep}, \textit{EvoMol-RL} significantly outperforms the baseline (\textit{EvoMol}) in pre-filtering realism percentage for both ECFP$_0$ ($p$-value $< 0.002$) 
%, $H_0$ is rejected)
and ECFP$_2$ ($p$-value $ < 0.002$).
%, $H_0$ is rejected).} 
When comparing the two fingerprint granularities (ECFP$_0$ and ECFP$_2$), \textit{EvoMol-RL} significantly delivers higher realism  when rewards are computed with ECFP$_2$ rather than ECFP$_0$ ($p$-value $< 0.002$) %, $H_0$ is rejected) 
(See Table \ref{tab: Results}). Since the larger diameter ($d=2$) captures chemical context beyond the focal atom that ECFP$_0$ ignores, this richer and more discriminative feature set gives the \textit{RL} policy clearer signals about which mutations are chemically plausible. This allows it to converge more quickly and to surpass the ECFP$_0$ variant across the entire $\varepsilon$ range, with the largest margin at $\varepsilon=0.1$.

Furthermore, Fig.~\ref{fig:PctRealismstep} shows that the new \textit{RL} module rapidly identifies effective mutations within the first $\approx 20$ steps, achieving a success rate that quickly surpasses and consistently outperforms the baseline.

Finally, the simultaneous drop in novelty and rise in realism (See Table~\ref{tab: Results}) is an expected indicator of effective reinforcement learning. Once the policy discovers mutation patterns that consistently produce realistic structures, it shifts from extensive exploration to targeted exploitation, revisiting those chemotypes more often. This deliberate focus lowers novelty but confirms that the agent is focusing on reliable, chemically plausible regions of the search space.

\section{Conclusion}\label{Conclusion}
In this article, the concept of context linked to an action on a molecular graph was introduced, paving the way for the use of reinforcement learning in molecular generation.
The main results of this article is the experimental demonstration that the addition of reinforcement learning for mutation selection in an evolutionary algorithm, here \textit{EvoMol}, allows the algorithm to be better guided towards solutions satisfying an objective function, the realism of molecules in this study. For this crucial objective for chemists, a significant improvements regardless of the methods and their hyperparameters was measured. Furthermore, the method converges rapidly and exhibits low variability between the different executions for a given set of parameters. It has also been shown that a larger context, using ECFP$_2$, allows a better policy learning.

The definition of the molecular context and this first application make it possible to consider direct extensions of this work in the future. Other reinforcement algorithms, such as EXP3 \cite{neu2015explore} to name just one, could be studied. In this study, the exploration capabilities of the original algorithm were inhibited in order to measure the specific contribution of reinforcement learning to mutation selection. A more exhaustive study, using all the mechanisms and varying the parameters, should be carried out. Other objective functions could also be tested, especially if their computational cost is very high, as in this case avoiding unnecessary evaluations may be critical. The next step in this work could be to look at deferred rewards, by letting the algorithm perform several actions before evaluating the molecule and calculating the reward to be assigned to each action. This extension is important in order to consider hard optimization problems requiring temporary non-improving our invalid solutions.

\bibliographystyle{IEEEtran}
\bibliography{UA_theo}
\end{document}